# A SURVEY OF ACTIVITY RECOGNITION AND UNDERSTANDING THE BEHAVIOUR IN VIDEO SURVEILLANCE

A.R.Revathi[1] and Dhananjay Kumar[2]

[1]Research Scholar, Department of Information Technology, Anna University, MIT campus

revathirajendran2010@gmail.com

[2]Associate Professor, Department of Information Technology, Anna University, MIT campus

dhananjayks@gmail.com

## ABSTRACT

*This paper presents a review of human activity recognition and behaviour understanding in video sequence. The key objective of this paper is to provide a general review on the overall process of a surveillance system used in the current trend. Visual surveillance system is directed on automatic identification of events of interest, especially on tracking and classification of moving objects. The processing step of the video surveillance system includes the following stages: Surrounding model, object representation, object tracking, activity recognition and behaviour understanding. It describes techniques that use to define a general set of activities that are applicable to a wide range of scenes and environments in video sequence.*

## 1. INTRODUCTION

As an active research topic in computer visions are the dynamic scenes detection, classifying object, tracking and recognizing activity and description of behaviour. Visual surveillance strategies have long been in use to gather information and to monitor people, events and activities. Video surveillance works as to detect moving object [1], [3], [6], [7], classify [8], [10] the detected object track [11], [13] them through the sequence of images and analysis the behaviours. Visual surveillance technologies [22], CCD cameras, thermal cameras and night vision device are the three most widely used devices in the visual surveillance market. The main goal of visual surveillance is not only to monitor, but also to automate the entire surveillance task. The goal of visual surveillance is to develop intelligent visual surveillance to replace the traditional passive video surveillance that is proving in effective as the numbers of cameras exceed the capability of human operators to monitor them. The automated surveillance systems can be implemented for both offline like storing the video sequence and to analyse the information in that sequence. But now days online surveillance system is very much needful in all public and private sectors due to predict and avoid unwanted movements, terrorist activities in those areas. It is helpful for traffic monitoring, transport networks, traffic flow analysis, understanding of human activity [20], [16], [15], home nursing, monitoring of endangered species, and observation of people and vehicles within a busy environment along many others to prevent theft and robbery. Some of the areas where video surveillance system place a major role in many application are 1) for military security 2)patrolling of country borders 3)extracting statistics for sport activities 4)surveillance of forests for fire detection 5) patrolling of highways and railway for accident detection.

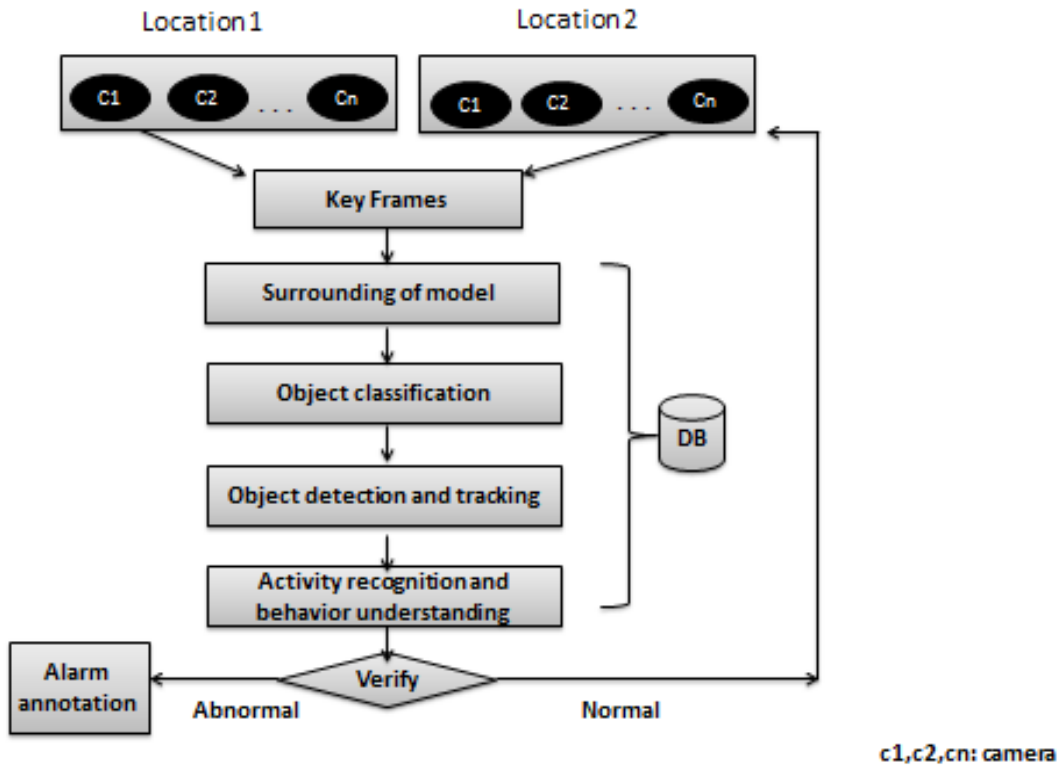

**Figure 1: Overview of an automated visual surveillance.**

In a video stream monitoring or analysing suspected anomalies can be divided in to three categories, which are single person with single object interaction, multiple people with multiple objects, and group behaviour[18], [13]. This paper focused to visual surveillance in the direction of surrounding of model, objects segmentation, classification, tracking and finally reached to behaviour understanding and action recognition. Many papers have been introducing various new concepts in this way. This paper reviews general strategies of various levels involved in video surveillance and analysis to challenges and feasibilities for combining object tracking, motion analysis, behaviour analysis for human subject understanding and behaviour understanding. The inspiration of writing and survey paper on this topic for a domain specific application is to review and gain insight in video surveillance. Surrounding of model includes object detection and object segmentation. It could be carried out by different approaches includes frame to frame difference, background subtraction and motion analysis using optical flow techniques. This becomes very popular due to robustness to noise, shadow and changing of light conditions.

## 2. SURROUNDING OF MODEL

Surrounding modelling is also known as Background modelling .It is currently used to detect moving objects in video acquired from static cameras. Numerous statistical methods have been developed over the recent years. The aim of this paper is to provide an extended and updated survey of the recent [23], [24], [25] researches which concern statistical background modelling. Murshed, M was proposed an edge segment based statistical background modelling [1] algorithm and a moving edge detection framework for the detection of moving objects. This paper actually focused about various methods of background modelling like traditional pixel based, edge pixel based and edge segment based approaches. He proposed this background modelling for natural image sequence with presence of illumination variation and noise. Yun Chu Zhang was analysed the background mechanism using GMM [2] model. Here this model updates new strategy which weighs the model adaptability and motion segmentation accuracy. But these works not focus dynamic moves in frame sequence. Wei Zhou [5] was proposed the dynamic background subtraction using spatial colour binary patterns. In addition to a refined model is designed to refine contour of moving objects. This method improves the accuracy of subtracting and detecting moving objects in dynamic scenes with presence of data driven model.

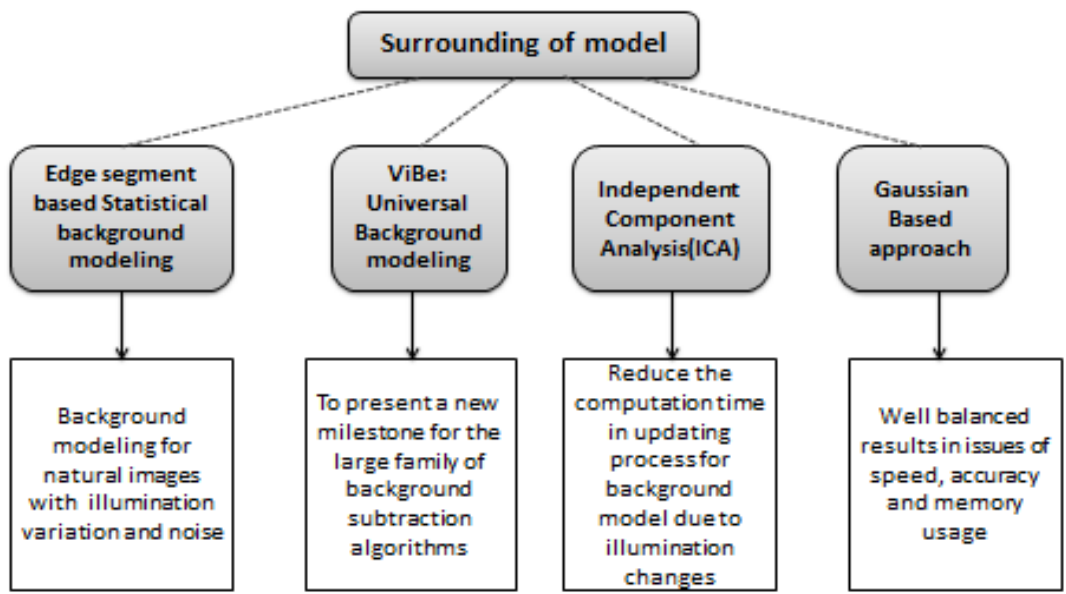

Figure 2: Overview of surrounding model.

Richard J.Radke [23] has written survey about image change detection algorithm with several challenge issues with solved by Stauffer and Grimson's background modelling with real times image. ViBe: A Universal Background Subtraction Algorithm [6] for Video Sequences paper presence a technique for motion detection which stores a set of value taken in the past in the same location or in the neighbourhood. It then compares this set to current pixel value in order to determine whether the pixel belongs to the background and to adopt the model which substitutes from the background model. There is a computation overhead in background model updating process due to illumination changes; Du-Ming Tsai [34] has proposed a fast background subtraction scheme using independent component analysis (ICA). In the training stage, an ICA model that directly measures the statistical independency based on the estimations of joint and marginal probability density functions from relative frequency distributions is first proposed. In the detection stage, the trained de-mixing vector is used to separate the foreground in a scene image with respect to the reference background image. M.Hedayati [35] suggested Gaussian-based approach is the best approach for real-time applications. He evaluated and compare five well know background like Median filtering, Approximate Median, Running Gaussian Average (RGA), Gaussian Mixture Modal (GMM) and Kernel Density Estimation (KDE). After compare these methods author focused there is not a perfect system at present, because the perfect system has to solve many problems such as bootstrapping, illumination changes and small movement in background. Gaussian based approached (RGA, GMM) give well balanced results in issues of speed, accuracy and memory usage for real time application.

**Table 1: Summary of surrounding model**

| Year | Author | Concepts |
|------|--------|----------|
| 2009 | ] Ddu-Ming Tsai, Shia-Chih Lai | a fast background subtraction scheme using independent component analysis (ICA)[34]. |
| 2010 | Murshed, M, Ramirez. A, Chae. O | An Edge Segment Based statistical background modelling [1]. |
| 2010 | M.Hedayati, Wan Mimi Diyana Wan Zaki, Aini Hussain | Gaussian-based approach for real-time applications [35] |
| 2011 | Yun Chu Zhang, Ru Min Zhang, Shi Jun Song | GMM Background and Covariance Estimation [2] |
| 2011 | Barnich.O, Van Droogenbroeck.M | Universal Background Subtraction for Video Sequences[6] |

# 3. OBJECT REPRESENTATION AND CLASSIFICATION

Object classification is a standard pattern recognition task. To track objects and analyse the behaviour, it is essential to correctly classify moving objects. There are two different categories of approaches for classifying moving objects like, shape based and motion based classification. Lun Zhang [7] has introduced a novel method for real time robust object classification in diverse camera viewing angles by applying appearance based techniques. A new descriptor, the multi block local binary pattern is used to capture the large scale structure in object appearance. An adaboost algorithm is introduced to select a subset of selected features as well as construct a different classifier for classification. However the above method has a overhead of the real time classification in the automated surveillance. To overcome real time classification, Hota, R.N [8] has proposed a online feature selection method which gives a good subset of features while the machine learns the classification task and use these selected features for object classification.

**Table 2: Summary of object representation and classification**

| Year | Author | Concept |
|------|--------|---------|
| 2007 | Lun Zhang, Li S.Z, Xiaotong Yuan, Shiming Xiang | Multi-block local binary pattern (MB-LBP), is proposed to capture the large-scale structures in object appearances [7]. |
| 2007 | Hota, R.N.  Venkoparao, V. Rajagopal, A | Shape Based Object Classification for Automated Video Surveillance with Feature Selection [8]. |
| 2012 | Robert Sorschag | A Flexible Object-of-Interest Annotation Framework for Online Video Portals. To present an automatic selection approach to support the use of different recognition strategies for different objects [9]. |

All the above methods have very little accuracy in the multi class objects. Yaniv Gurwicz [31] has proposed a Multiclass object classification for real-time video surveillance systems which introduce an approach for classifying objects in both low and high resolution images in real world scenarios. In this method have several features that jointly leverage the divisions between various classes and also provide a feature selection procedure based on the entropy gain.

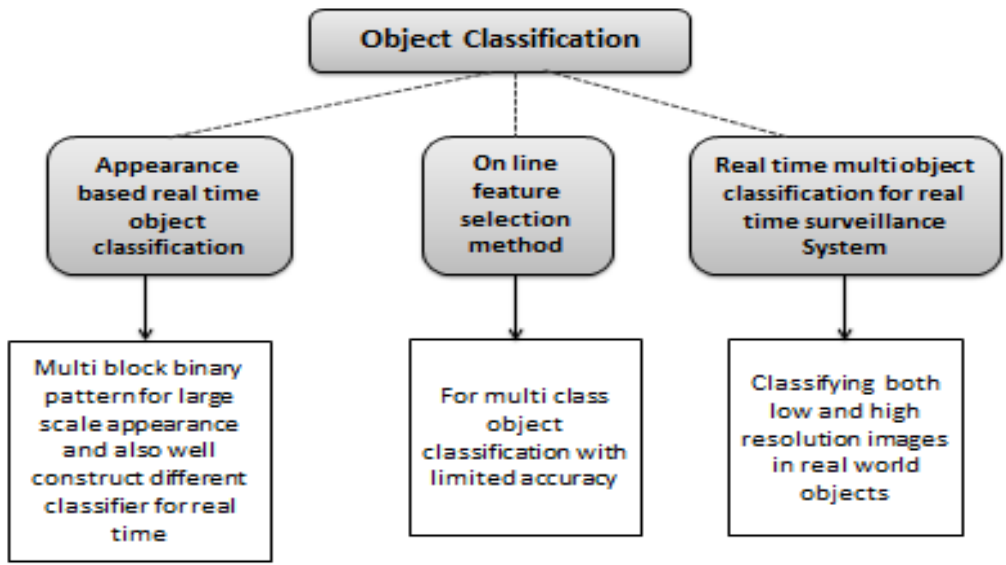

Figure 3:  Overview of Object classification.

Robert Sorschag [9] has proposed a Flexible object-of-interest Annotation framework for online video portals using content based analysis method. This framework slightly varies from other content based analysis method, since it possesses some outstanding features that offer good prospects for real video portal applications.

# 4. OBJECT TRACKING

The objective of video tracking is to associate target objects in consecutive video frames. The association can be especially difficult when the objects are moving fast relative to the frame rate. Blob tracking, kernel-based tracking, Contour tracking are some common target representation and localization algorithms. Ruolin Zhang [32] has proposed adaptive background subtraction about the video detecting and tracking moving object. He use median filter to achieve the background subtraction. This algorithm is used for both detecting and tracking moving objects in sequence of video. This algorithm never support for multi feature based object detection. Hong Lu and Hong Sheng Li [10] were introduced a new approach to detect and track the moving object. The affine motion model and the non-parameter distribution model are utilized to represent the object and then the motion region of the object is detected by background difference while Kalman filter estimating its affine motion in next frame. The author shows Experimental results and proof the new method can successfully track the object under such case as merging, splitting, scale variation and scene noise. The author Bayan [11] talks about adaptive mean shift for automated multi tracking. Mixture of Gaussian model extracted Foreground and then followed by shadow and noise removal to initialise the object trackers and also to make the system more efficient by decreasing the search area and the number of iterations to converge for the new location of the object. Trackers are automatically relaxed to solve the potential problems that may occur because of the changes in objects' size, shape, to handle occlusion-split between the tracked objects and to detect newly emerging objects as well as objects that leave the scene. Using a shadow removal method has increases the tracking accuracy.

Table 3: Summary of object tracking

| Year | Author | Concept |
|------|--------|---------|
| 2011 | Hong Lu, Hong Sheng Li, Lin Chai, Shu Min Fei, Guang Yun Liu | Multi-Feature Fusion Based Object Detecting and Tracking. Then the motion region of the object is detected by background difference while Kalman filter estimating its affine motion in next frame [10]. |
| 2011 | Blanco Adán, Carlos Roberto del, Jaureguizar Nuñez, Fernando, García Santos | Bayesian Visual Surveillance, a Model for Detecting and Tracking a variable number of moving objects [13]. |
| 2012 | Beyan, C. Temizel, A | Adaptive mean-shift for automated multi object tracking [11]. |
| 2012 | Jiu Xu, Chenyuan Zhang | Modified codebook foreground detection and particle filter [12]. |
| 2012 | Lin Lizhong , Liu Zhiguo , Liu Zhiguo | Intelligent video surveillance on tracking of moving targets [36]. |
| 2012 | Louis Kratz , Ko Nishino | Local spatiotemporal motion patterns and HMM for tracking pedestrians [37]. |

Jiu Xu [12] was proposed the foreground object detection process by the modified code book. The modified codebook foreground detection and particle filter are used for object tracking by detection for surveillance system. This code book was created by adding the orientation and magnitude of the block gradient, the codebook model contains not only information of color, but also the texture feature in order to further reduce noises and refine more entire foreground regions using modified code book. By calculate the local binary pattern of the edge of the foreground objects which could have a good performance in describing the shape and texture of the objects. But this will give only solution for one or two object with particle occlusion. Bayesian Visual Surveillance [13], a Model for Detecting and Tracking a variable number of moving objects paper was focus for many objects with occlusion. Video object detectors generate an unordered set of noisy, false, missing, split, and merged measurements that make extremely complex the tracking task. He was introduced a Bayesian Visual Surveillance Model is proposed that is able to manage undesirable measurements. Particularly, split and merged measurements are explicitly modelled by stochastic processes. Lin Lizhong has proposed [36] an intelligent video surveillance system mainly carries out research on techniques of detection and tracking of moving targets, which are very important for detection and understanding of abnormal behaviors, so the effect of moving detection directly affects the overall effect of video surveillance systems, moving detection of intelligent. Detection and tracking of moving targets in surveillance applications has greatly improved intelligence, accuracy and reliability of video surveillance and has greatly reduced the burden on staff. Intelligent analysis module is the core of the video surveillance system, which mainly includes the number of target identification and extraction of statistics. After encoding, compression, transmission and pre-processing of the video images are made, then the detection algorithm is used to extract key information, the next step is to facilitate the analysis of image understanding. Previous work on tracking is mostly object centric, based on the modeling of the motion and appearance of the target individual. Louis Kratz[37] has proposed a novel, space-time model of the crowd motion as a set of priors for tracking individuals in videos of extremely crowded scenes. Author begins by representing the crowd motion as a statistical model of spatially and temporally varying local spatiotemporal motion patterns. HMM is capable to accurately predict the changing motion of pedestrians by encoding the scene's underlying temporal dynamics, and this method is more accurate than those that do not model the space-time variations in the crowd motion.

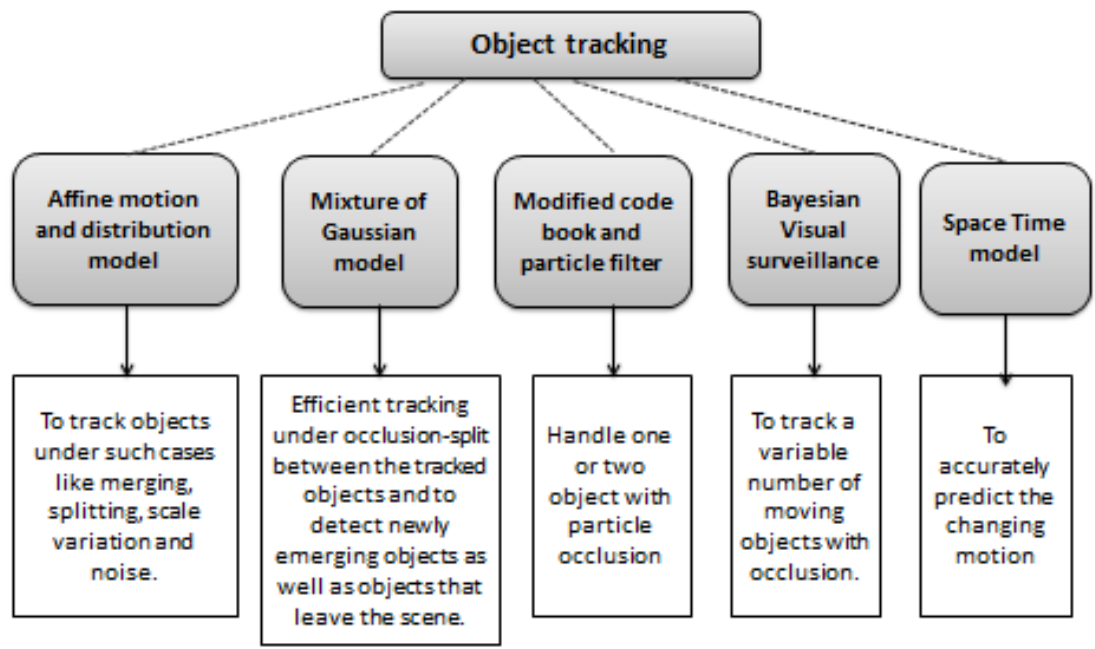

Figure 4: Overview of Object tracking.

## 5. DESCRIPTION AND BEHAVIOUR UNDERSTANDING

Human action recognition is a challenging zone of research because of its various potential applications in visual surveillance. Automatic human activity recognition in video using background modelling and spatio-temporal template matching based technique [28] can be proposed by Chandra Mani Sharm. This activity recognition processed based on the spatio-temporal template matching. The Motion history images are used to construct the Spatio-temporal templates and object shape information for different human activities in a video like walking, standing, bending, sleeping and jumping. This method accurately measure all the activities.

**Table 4: Description and behaviour understanding**

| Year | Author | Concept |
|------|--------|---------|
| 2008 | Tao Xiang and Shaogang Gong | Online LRT based behaviour recognition approach is advantageous over the commonly used Maximum Likelihood (ML) method in differentiating ambiguities among different behaviour classes observed online [30]. |
| 2011 | Chandra Mani Sharma, Alok Kr. Singh Kushwaha, Swati Nigam, Ashish Khare | Automatic human activity recognition in video using background modeling and spatio-temporal template matching based technique [28]. |
| 2011 | Chun-hao Wang, Yongjin Wang, Ling Guan | Event detection and recognition using histogram of oriented gradients and hidden markov models [27]. |
| 2011 | Cheng Chen; Heili, A.; Odobez, J.-M | Human behaviour analysis by joint temporal filtering [39]. |
| 2012 | Arnold Wiliem, Vamsi Madasu, Wageeh Boles and Prasad Yarlagadda | Context space model for detecting anomalous behaviour [38]. |

Chun-hao Wang [27] presents an approach for object detection and action recognition in video surveillance scenarios. This system utilizes a Histogram of Oriented Gradients (HOG) method for object detection, and a Hidden Markov Model (HMM) for capturing the temporal structure of the features. Decision making is based on the understanding of objects motion trajectory and the relationships between objects' movement and events. David Nicholas Olivieri proposed[29] such software based upon a spatio-temporal motion representation, called Motion Vector Flow Instance (MVFI) templates that capture relevant velocity information by extracting the dense optical flow from video sequences of human actions. Tao Xiang [30] was proposed novel framework is developed for automatic behaviour profiling and online anomaly sampling/detection without any manual labelling of the training data set. The framework consist various key components: A compact and effective behaviour representation method is developed based on discrete-scene event detection. The similarity between behaviour patterns are measured based on modelling each pattern using a Dynamic Bayesian Network (DBN).

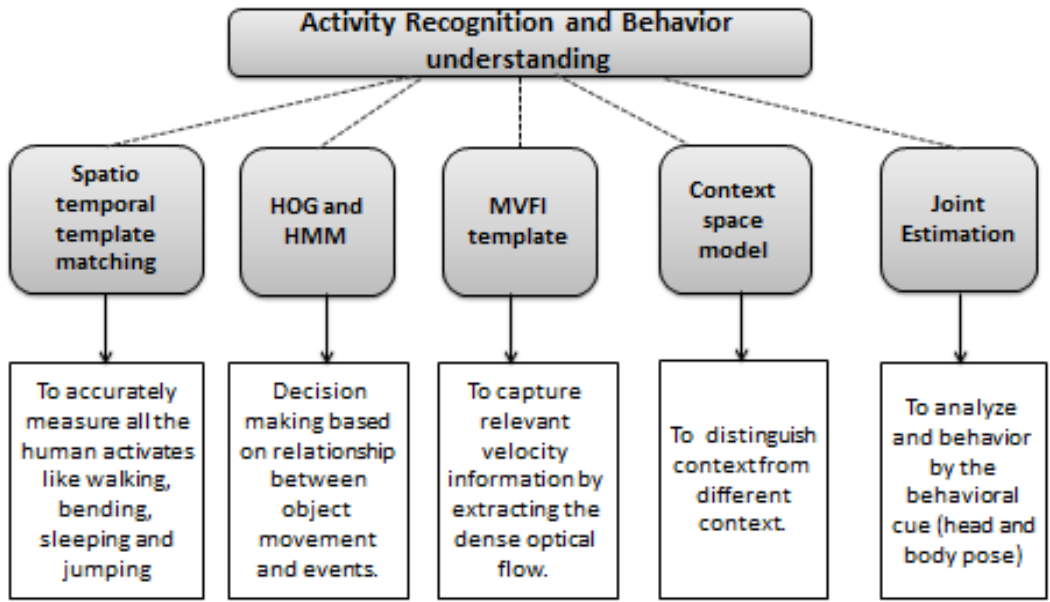

Figure 5: Overview of activity recognition and Behavior understanding.

The natural grouping of behaviour patterns is discovered through a novel spectral clustering algorithm with unsupervised model selection and feature selection on the eigenvectors of a normalized affinity matrix. A composite generative behaviour model is constructed that is capable of generalizing from a small training set to accommodate variations in unseen normal behaviour patterns. A runtime accumulative anomaly measure is introduced to detect abnormal behaviour, whereas normal behaviour patterns are recognized when sufficient visual evidence has become available based on an online Likelihood Ratio Test (LRT) method. Arnold Wiliem [38] has proposed a context space model for detecting Anomalous Behaviour in video sequence. Context information is very much useful to create a detection system. Whereas in existing surveillance, the use of contextual information is limited in automatic anomalous human behaviour detection approaches. This context space model provides guidelines for the system designers to select information which can be used to describe context and also enables a system to distinguish between different contexts. Context space is defined as an n-dimensional information space formed by context parameters selected from the contextual information as its bases. Cheng Chen and Alexandre Heili [39] has proposed a method for analysis and understanding behaviour by joint estimation of human behavioural cue(like head and body pose) from video frames. In the frame analysis the head is localized, head and body features are extracted, and their likelihoods under different poses are evaluated. The author made two contributions first level is a head localization method which reliably localizes the head of a person from a human detection bounding box, and a head pose classifier estimating the head pose from the localization output; second level is a framework for the joint estimation of body position, body pose and head pose, relying on the soft coupling between these cues.

## 6. DATA FUSION

Data fusion is important for occlusion handling and continuous tracking. Occlusions remove the direct correspondence between visible areas of objects and the objects themselves by introducing ambiguity in the interpretation of the shape of the occluded object. Dockstader et al. [18] use a Bayesian network to fuse 2-D state vectors acquired from various image sequences to obtain a 3-D state vector. Collins et al. [19] design an algorithm that obtains an integrated representation of an entire scene by fusing information from every camera into a 3-D geometric coordinate system. Kettnaker et al. [17] synthesize the tracking results of different cameras to obtain an integrated trajectory.

**Table 5: Summary of widely used techniques for visual surveillance**

| Year | Author | Concepts |
|------|--------|----------|
| 2007 | Lun Zhang, Li S.Z, Xiaotong Yuan, Shiming Xiang | Object Classification in Video Surveillance Based on Appearance Learning [7]. |
| 2007 | Hota.R.N.  Venkoparao.V  Rajagopal.A | Shape Based Object Classification for Automated Video Surveillance with Feature Selection [8]. |
| 2008 | Tao Xiang and Shaogang Gong | Video Behavior Profiling for Anomaly Detection [30] |
| 2009 | Ddu-Ming Tsai, Shia-Chih Lai | A fast background subtraction scheme using independent component analysis (ICA)[34]. |
| 2010 | Murshed, M, Ramirez. A, Chae. O | An Edge Segment Based Moving Object Detection Approach [1]. |
| 2010 | M.Hedayati, Wan Mimi Diyana Wan Zaki, Aini Hussain | Gaussian-based approach for real-time applications [35] |
| 2011 | Yun Chu Zhang, Ru Min Zhang, Shi Jun Song | Research on GMM Background Modeling and Covariance Estimation [2]. |
| 2011 | Barnich.O,Van roogenbroeck.M | ViBe: A Universal Background Subtraction Algorithm for Video Sequences [6]. |
| 2011 | Hong Lu, Hong Sheng Li, Lin Chai, Shu Min Fei, Guang Yun Liu | Multi-Feature Fusion Based Object Detecting and Tracking [10]. |
| 2011 | Blanco Adán, Carlos Roberto del, Jaureguizar Nuñez, Fernando, García Santos | Bayesian Visual Surveillance, a Model for Detecting and Tracking a variable number of moving objects [13]. |
| 2011 | Chandra Mani Sharma,  Alok  Kr. Singh Kushwaha,  Swati  Nigam, Ashish Khare | Automatic human activity recognition in video using background modeling and spatio-temporal template matching based technique [28]. |
| 2011 | Chun-hao  Wang,  Yongjin  Wang, Ling Guan | Event detection and recognition using histogram of oriented gradients and hidden markov models [27]. |
| 2011 | Cheng Chen; Heili, A.; Odobez, J.-M | Human behaviour analysis by joint temporal filtering [39]. |
| 2012 | Robert Sorschag | A Flexible Object-of-Interest Annotation Framework for |
| | | Online Video Portals [9]. |
| 2012 | Beyan, C.  Temizel, A | Adaptive mean-shift for automated multi object tracking [11]. |
| 2012 | Jiu Xu, Chenyuan Zhang | Modified codebook foreground detection and particle filter [12]. |
| 2012 | Lin Lizhong , Liu Zhiguo , Liu Zhiguo | Intelligent video surveillance on tracking of moving targets[36]. |
| 2012 | Louis Kratz , Ko Nishino | Local spatiotemporal motion patterns and HMM for tracking pedestrians [37]. |
| 2012 | Arnold Wiliem, Vamsi Madasu, Wageeh Boles and Prasad Yarlagadda | Context space model for detecting anomalous behaviour [38]. |

# 7. CONCLUSION

Most current automated video surveillance systems can process video sequence and perform almost all key low-level functions, such as motion detection and segmentation, object tracking, and object classification with good accuracy. But technical interest in video surveillance has moved from such low-level functions to more complex scene analysis to detect human and/or other object behaviors, i.e., patterns of activities or events, for standoff threat detection and prevention. Existing behavior analysis systems focus on the predefined behaviors, e.g., to combine the results of an automated video surveillance system with spatiotemporal reasoning about each object relative to the key background regions and other objects in the scene. Advanced behavior analysis systems have begun to exploit the capability to automatically capture and learn new behaviors by pattern matching, and further present the behavior to the specialists for confirmation. This paper reviews and exploits developments and general strategies of stages involved in video surveillance and analyses the challenges and feasibility for combining object tracking, motion analysis, behavior analysis, and biometrics for stand-off human subject identification and behavior understanding. Behavior understanding and activity recognition using visual surveillance involves the most advanced and complex researches in image processing, computer vision, and artificial intelligence. There were many diverse methods have been used while approaching this challenge; and they varied and depended on the required speed, the scope of application, and resource availability, etc.

A.R.Revathi received the Master of Technology in Computer Science and Engineering from SASTRA University in 2002 and received Bachelor of Engineering in Electronic and Communication and Engineering from Bharathidasan University in 2000.She is currently working as Assistant Professor in Department of Information Technology at SRM Valliammai Engineering College. Her research interests are mainly focused on motion detection, human detection and recognition, and computer vision.

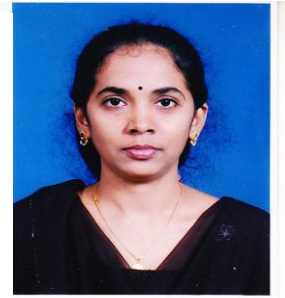

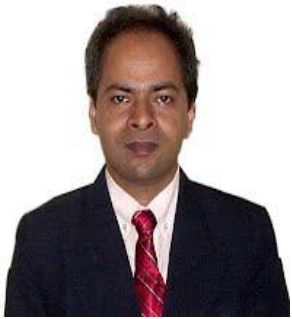

Dr.Dhananjay Kumar   received Doctor of Philosophy in Information & Communication Engineering from Anna University in 2009. He received the Master of Engineering in Industrial Electronics Engineering from Baroda University in 1999 and Master of Technology in Communication Engineering from Pondicherry University in 2001. His primary job at the department of Information Technology, Anna University Chennai involves teaching and research. His technical interest includes mobile computing & communication, multimedia systems, and signal processing.